\documentclass[sigconf,nocopyrightspace]{acmart}

\usepackage{colortbl}
\usepackage{multirow}
\usepackage{makecell}
\usepackage{booktabs}
\usepackage{colortbl}
\usepackage{xcolor}
\usepackage{graphicx}
\usepackage{caption}
\usepackage{subcaption}
\usepackage{bm}
\usepackage{algorithm}    
\usepackage{algpseudocode}

\begin{document}

\title{ReaLM: Residual Quantization Bridging Knowledge Graph Embeddings and Large Language Models}

\author{Wenbin Guo}
\email{wenff@tju.edu.cn}
\affiliation{%
  \institution{Tianjin University}
  \city{Tianjin}
  \country{China}
}

\author{Xin Wang}
\email{wangx@tju.edu.cn}
\affiliation{%
  \institution{Tianjin University}
  \city{Tianjin}
  \country{China}
}

\author{Jiaoyan Chen}
\email{jiaoyan.chen@manchester.ac.uk}
\affiliation{%
  \institution{The University of Manchester}
  \city{Manchester}
  \country{United Kingdom}
}

\author{Lingbing Guo}
\email{lbguo@tju.edu.cn}
\affiliation{%
  \institution{Tianjin University}
  \city{Tianjin}
  \country{Chile}
}

\author{Zhao Li}
\email{lizh@tju.edu.cn}
\affiliation{%
  \institution{Tianjin University}
  \city{Tianjin}
  \country{China}
}

\author{Zirui Chen}
\email{zrchen@tju.edu.cn}
\affiliation{%
  \institution{Tianjin University}
  \city{Tianjin}
  \country{China}
}

\settopmatter{printacmref=false}
\renewcommand\footnotetextcopyrightpermission[1]{}

\begin{abstract}
  Large Language Models (LLMs) have recently emerged as a powerful paradigm for Knowledge Graph Completion (KGC), offering strong reasoning and generalization capabilities beyond traditional embedding-based approaches. However, existing LLM-based methods often struggle to fully exploit structured semantic representations, as the continuous embedding space of pretrained KG models is fundamentally misaligned with the discrete token space of LLMs. This discrepancy hinders effective semantic transfer and limits their performance.
  To address this challenge, we propose ReaLM, a novel and effective framework that \textit{bridges the gap between KG embeddings and LLM tokenization} through the mechanism of \textit{residual vector quantization}. ReaLM discretizes pretrained KG embeddings into compact code sequences and integrates them as learnable tokens within the LLM vocabulary, enabling seamless fusion of symbolic and contextual knowledge. Furthermore, we incorporate ontology-guided class constraints to enforce semantic consistency, refining entity predictions based on class-level compatibility.
  Extensive experiments on two widely used benchmark datasets demonstrate that ReaLM achieves state-of-the-art performance, confirming its effectiveness in \textit{aligning structured knowledge with large-scale language models}. The implementation is publicly available at https://anonymous.4open.science/r/ReaLM-D270. 
\end{abstract}

\maketitle

\section{Introduction}
Knowledge Graphs (KGs) have become indispensable infrastructures for organizing and representing structured knowledge across diverse domains, including search engines \cite{efficient}, recommender systems \cite{learning}, and biomedical informatics \cite{beyond}. By encoding factual information into triples of entities and relations, KGs provide a machine-interpretable foundation for reasoning and knowledge-driven applications. Yet, real-world KGs are inherently incomplete, as many valid facts remain unobserved due to the high cost of manual curation and the evolving nature of knowledge \cite{ConvD}. Addressing this incompleteness through Knowledge Graph Completion (KGC), which is a task of inferring missing links, has therefore become a central research problem for enhancing the coverage and reliability of knowledge-centric systems \cite{TransE}.  

\begin{figure}[h]
  \centering
  \includegraphics[width=\linewidth]{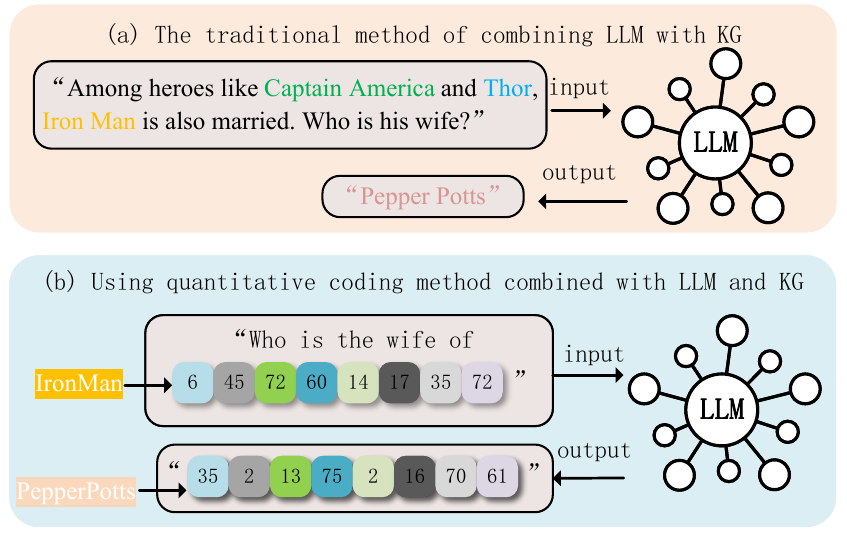}
  \caption{Motivation of ReaLM: replacing text-based entity representations with quantized code sequences for explicit and efficient LLM–KG integration.}
  \label{fig:Motivation}
\end{figure}

Recent advances in large language models (LLMs) have inspired attempts to leverage their reasoning capabilities for KGC, yielding promising results in triple classification and link prediction tasks. \textbf{Nevertheless, a fundamental challenge remains: the intrinsic mismatch between continuous KG entity embedding and the discrete token space of LLMs} \cite{KoPA}. 
When a KG is directly converted into textual form and fed into an LLM, the generated results are expressed in natural language tokens rather than the exact entities defined in the KG. Consequently, these outputs cannot be directly aligned with the entity embeddings used in the original graph, leading to inconsistencies and ambiguity in entity mapping. A straightforward alternative is to incorporate each KG entity as a unique token within the LLM vocabulary, thereby enabling explicit generation of KG entities. However, this approach introduces substantial challenges. Unlike natural language vocabularies, which are fixed and relatively compact, knowledge graphs generally encompass a vast and continuously evolving collection of entities, often reaching tens or even hundreds of thousands in number, and in some cases surpassing the token vocabulary size of standard large language models. For instance, the LLaMA-2 model employs a vocabulary of 32,000 tokens, while the WN18RR dataset alone contains 40,943 distinct entities \cite{MKGL}.
Expanding the LLM vocabulary to include all these entities would dramatically increase the token space, leading to excessive memory consumption, slower training, and unstable inference. As a result, generative LLM-based approaches struggle to produce valid and consistent KG entities directly, limiting their effectiveness and scalability in practical KGC scenarios.

Several recent efforts have sought to bridge the gap between LLMs and KGs for completion tasks. For example, MKGL converts KG triples into textual corpora and fine-tunes LLMs to directly generate entities, but this approach often yields semantically related yet invalid outputs that deviate from the exact entity set \cite{MKGL}. KoPA adopts a different strategy by employing a prefix adapter that projects entity and relation embeddings into the same dimensional space as tokens, enabling LLMs to perform triple classification. However, such a direct dimensional transformation neither supports ranking-based tasks nor fully captures the rich semantics inherent to KGs \cite{KoPA}. More recently, SSQR introduces a novel quantization-based approach: it reconstructs structural information and employs semantic distillation losses to map entity embeddings into a discrete codebook, which is then partially integrated into the LLM tokenizer \cite{SSQR}. While this method achieves competitive performance in KGC, it requires the careful joint optimization of structural reconstruction, semantic distillation, and quantization objectives, which is an intricate process highly prone to imbalance during training.  

To address these limitations, we propose \textbf{ReaLM}, a unified framework that seamlessly integrates \textbf{semantics of pretrained KG embeddings} into LLMs while preserving both \textbf{semantic fidelity and computational efficiency}. 
As illustrated in Figure~\ref{fig:Motivation}, traditional approaches typically convert triples into textual descriptions and rely on the LLM to generate entity names in natural language. However, such text-based formulations often fail to align generated tokens with the \textbf{discrete entities defined in the KG}. In contrast, ReaLM adopts a \textbf{quantitative coding paradigm}, where KG entities are represented by sequences of compact discrete codes derived through \textbf{residual vector quantization}. This encoding strategy preserves the \textbf{relational structure and semantic nuances} of pretrained embeddings while ensuring compatibility with the limited token space of LLM, thereby avoiding uncontrolled vocabulary expansion. The quantized codes are incorporated as new tokens within the LLM vocabulary and jointly optimized with its internal parameters, enabling the model to \textbf{internalize KG semantics without sacrificing linguistic fluency}.
Furthermore, ReaLM integrates \textbf{ontology-guided class constraints}, ensuring that entity predictions remain consistent with their corresponding classes, which enhances both \textbf{semantic correctness}. By combining residual quantization with ontology-based refinement, ReaLM effectively addresses the challenges of semantic misalignment, vocabulary scalability, and generative precision. Our main contributions are summarized as follows:  
\begin{itemize}
    \item We introduce a residual vector quantization framework that transforms high-dimensional continuous KG embeddings into compact discrete code sequences, enabling seamless integration with LLM token spaces while preserving rich relational and semantic information.  
    \item We incorporate ontology-based class constraints to enforce semantic consistency between predicted entities and their corresponding classes, thereby enhancing both accuracy in KGC.  
    \item We demonstrate the effectiveness of our approach through extensive experiments on benchmark datasets, including FB15K237 and WN18RR, where ReaLM significantly outperforms existing LLM-based and traditional KG completion methods. 
\end{itemize}

\section{Related Work}

\subsection{Traditional Embedding-based KGC Models}

Traditional KGC models typically adopt geometric embedding approaches, assuming relational consistency, where relations imply specific transformations that align entity vectors \cite{MSHE}. Early methods, such as TransE \cite{TransE}, treat relations as translations between entity embeddings, while subsequent models enhance representational capacity through more sophisticated operator spaces, including hyperplanes \cite{TransH}, relational mapping spaces \cite{TransD}, Riemannian spheres \cite{5E}, multi-mapping spaces \cite{TransM}, rotational spaces \cite{RotatE}, and convolutional spaces \cite{ConvE,ConvKB,ConvR}.
The integration of attention mechanisms and deep architectures has markedly advanced KGC, allowing models to capture intricate relational dependencies and contextual patterns. Techniques such as relation-aware attention in RelAtt \cite{RelAtt}, path-aggregation strategies in HRAN \cite{HRAN}, and Transformer-based frameworks like KGT5 and HittER \cite{KGT5,HittER} enable scalable, context-sensitive representation learning, effectively modeling higher-order interactions within large-scale KGs.
Nevertheless, embedding-based approaches remain inherently limited, as they often fail to distinguish semantically similar yet class-inconsistent entities, that is, entities that are close in the embedding space but belong to different ontological classes within the knowledge graph. This limitation frequently results in semantically plausible but class-inconsistent predictions and weak generalization to unseen entities \cite{summar20241}. 

\subsection{Quantization-based and Parameter-efficient KGC Models}

Quantization in KG embedding is a pivotal technique that involves the transformation of high-dimensional continuous entity representations into a discrete or compact code space. This process is primarily aimed at reducing the memory footprint and computational cost associated with traditional embedding methods, while retaining the essential semantic information inherent in the entities and their relationships \cite{Li2021, Zhu2022}. Recent models have leveraged this idea to scale embeddings to large graphs efficiently. NodePiece \cite{Galkin2022} constructs a fixed-size entity vocabulary by selecting a subset of entities as anchors and representing each entity through its distances to these anchors. EARL \cite{Chen2023} incorporates relation information to improve anchor matching, and RandomEQ \cite{Li2023} simplifies the process through random selection of anchors and relations, achieving competitive results with minimal supervision.

These traditional quantization-based and parameter-efficient methods primarily focus on directly optimizing embeddings for reasoning, without considering the integration with LLMs, which could potentially enhance their performance by leveraging the contextual understanding and reasoning capabilities of LLMs. 

\subsection{LLM-Based KGC Models}

With the rapid advancement of LLMs, increasing efforts have been devoted to bridging the gap between the structured knowledge of KGs and the generative reasoning capabilities of LLMs. Pre-trained model approaches such as KG-BERT \cite{KG-BERT} reformulate triples into textual sequences and fine-tune BERT to classify their plausibility. While effective, they are constrained by their encoder-only architecture and struggle to seamlessly integrate the implicit knowledge of Pre-trained models with the explicit semantics of KGs \cite{KnowC}. In contrast, recent LLM-based KGC models exploit the generative and reasoning strengths of decoder-based architectures, either by converting structured KG information into textual inputs—often via zero-shot reasoning or instruction tuning—or by aligning KG embeddings with the token space of LLMs for tighter integration. Representative examples include MuKDC, which prompts LLMs with subgraph-based questions to infer missing triples \cite{MuKDC-1}, and KLR-KGC, which enhances reasoning through analogical and subgraph retrieval \cite{KLR-KGC}. Other methods, such as TEA-GLM, treat LLMs as zero-shot learners for graph machine learning tasks \cite{TEA-GLM}, while KGEditor encodes KG paths as chain-of-thought prompts to support multi-hop reasoning \cite{KGEditor}. MKGL converts KG triples into textual corpora and fine-tunes LLMs to generate entities directly, but often produces semantically related yet invalid outputs outside the predefined entity set \cite{MKGL}; KoPA introduces a prefix adapter to project entity and relation embeddings into the LLM token space, enabling effective triple classification but falling short in ranking tasks and semantic fidelity \cite{KoPA}; and SSQR reconstructs structural information and distills semantics into a discrete codebook partially integrated into the LLM tokenizer, though its joint optimization of reconstruction, distillation, and quantization introduces training instability \cite{SSQR}. Overall, while LLM-based methods significantly extend the reasoning capacity of KGC beyond traditional embedding models, they remain challenged by the need to generate predictions faithful to the closed KG vocabulary, maintain semantic consistency across entity classes, and ensure stable integration of structured embeddings with discrete token-based models.  

\section{Preliminaries}

Let $\mathcal{E}$ and $\mathcal{R}$ denote the sets of entities and relations in a KG, respectively, where each triple $(h, r, t) \in \mathcal{F} \subseteq \mathcal{E} \times \mathcal{R} \times \mathcal{E}$ represents a relation $r$ linking head entity $h$ to tail entity $t$. Each entity $e \in \mathcal{E}$ may also be associated with a class $y(e) \in \mathcal{Y}$. In the KGC task, the objective is to predict a missing entity in an incomplete triple, e.g., $(h, r, ?)$ or $(?, r, t)$, by ranking candidate entities based on a scoring function $f(h, r, t')$. When ontology knowledge is available, predictions are further constrained to ensure class consistency, accepting $\hat{t}$ only if $y(\hat{t}) = \hat{y}$. 

To enhance representation efficiency, continuous embeddings $\bm{e}$ capturing semantic and relational information can be quantized into discrete latent codes $[i_1, \dots, i_S]$, thereby enabling compact storage and efficient computation. In LLM-based frameworks, triples are reformulated as textual sequences, allowing the model to either generate the missing entity ID or evaluate the plausibility of candidate triples, bridging symbolic reasoning in KGs with the generative capability of LLMs.

\section{Methodology}

\begin{figure*}[h]
  \centering
  \includegraphics[width=\textwidth]{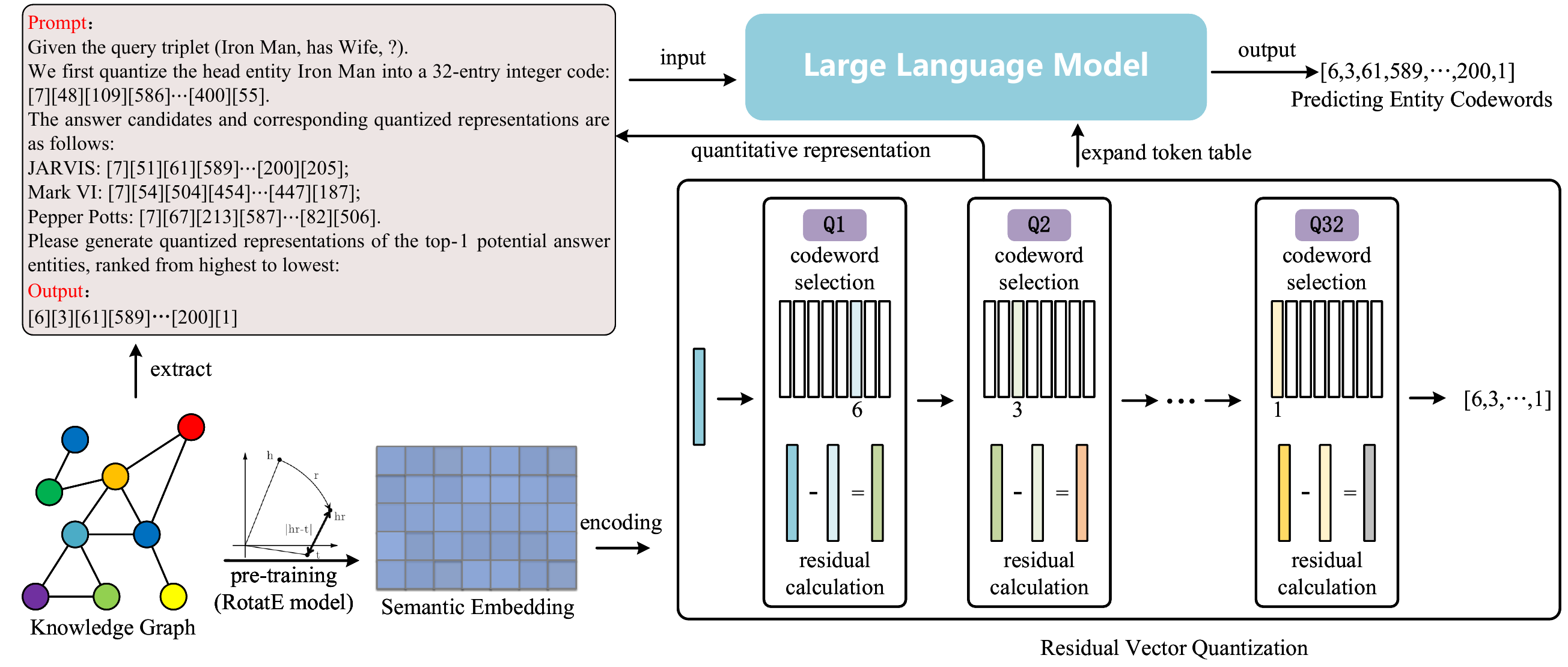}
  \caption{The overall framework of ReaLM.}
\end{figure*}

ReaLM integrates KG semantics into LLMs through a structured, multi-stage process. It first encodes entities into embeddings, then discretizes these embeddings via residual vector quantization to produce compact token representations. These tokens are incorporated into the LLM vocabulary for fine-tuning, while ontology-guided class information enforces semantic consistency during prediction. This design ensures that structured knowledge is preserved and effectively leveraged in the generative model.  

\subsection{Knowledge Graph Semantic Extraction}

A key challenge in integrating KGs with LLMs lies in extracting semantic representations that are both structurally informative and readily transferable to downstream tasks. While numerous embedding methods have been proposed for KGs, such as TransE, DistMult, and ComplEx, our preliminary evaluation demonstrates that \textbf{RotatE \cite{RotatE} consistently achieves superior performance} in capturing relational patterns crucial for KGC, making it particularly suitable as the semantic backbone for our framework.  

Formally, given a triple $(h, r, t) \in \mathcal{F}$, RotatE embeds entities $\bm{e}_h, \bm{e}_t \in \mathbb{C}^d$ and relation $\bm{e}_r \in \mathbb{C}^d$ in the complex vector space, where relations are parameterized as rotations. The scoring function is defined as: 
\begin{equation}
  f(h, r, t) = -\|\bm{e}_h \circ \bm{e}_r - \bm{e}_t\| 
\end{equation}
where $d$ representing the dimension of embedding, and $\circ$ represents element-wise (Hadamard) multiplication.

To obtain pretrained semantic embeddings, we train RotatE on the target KG until convergence using standard negative sampling and margin-based ranking loss. The objective is:

\begin{equation}
    \mathcal{L}_{\text{RotatE}} =
    - \sum_{(h,r,t)\in\mathcal{F}} \log\sigma\!\bigl(\gamma - f(h,r,t)\bigr)
    - \sum_{(h',r,t')\in\mathcal{F}'} \log\sigma\!\bigl(f(h',r,t')-\gamma\bigr)
\end{equation}
where $\mathcal{F}'$ is the set of negative samples constructed by corrupting the triples, $\gamma$ is a margin hyper-parameter, and $\sigma(\cdot)$ is the sigmoid function.

Through this pretraining process, we obtain high-quality entity embeddings $\{\bm{e} \in \mathbb{C}^d \mid e \in \mathcal{E}\}$, which encode both structural connectivity and semantic consistency. These embeddings serve as the \textbf{foundation for our residual vector quantization module}, ensuring that the subsequent quantization preserves rich semantic information while bridging the gap to LLM tokenization.

\subsection{Residual Vector Quantization of Semantic Embeddings}

Given a pretrained entity embedding $\bm{e} \in \mathbb{C}^d$ obtained from the RotatE model, our goal is to transform it into a discrete representation while minimizing information loss. To this end, we employ \textbf{Residual Vector Quantization}, which approximates $\bm{e}$ through a sequence of residual refinements across $S$ stages.  

Each stage $s \in \{1, \dots, S\}$ is associated with an independent codebook 
$\mathcal{C}_s = \{\mathbf{c}_{s,1}, \dots, \mathbf{c}_{s,K}\}$, 
where $K$ is the number of codewords and each $\mathbf{c}_{s,i} \in \mathbb{R}^{2d}$. 
Here, the original complex-valued RotatE embedding $\bm{e} \in \mathbb{C}^d$ is mapped to a real-valued vector 
$\bm{e}_\text{real} \in \mathbb{R}^{2d}$ by concatenating its real and imaginary parts, 
so that it can be directly processed by the real-valued residual vector quantization. 
At stage $s$, the residual vector $\mathbf{res}_{s-1}$ (with $\mathbf{res}_0 = \mathbf{e}_\text{real}$) 
is quantized by selecting the nearest codeword:

\begin{equation}
    i = \arg\min_{j} \|\mathbf{res}_{s-1} - \mathbf{c}_{s,j}\|, \quad
\mathbf{res}_s = \mathbf{res}_{s-1} - \mathbf{c}_{s,i}
\end{equation}
where $i$ denotes the index of the selected codeword at stage $s$. 
The operator $\arg\min_j$ returns the index $j$ that minimizes the distance 
between the residual vector $\mathbf{res}_{s-1}$ and the candidate codeword $\mathbf{c}_{s,j}$. The quantized embedding is then given by the cumulative approximation:  

\begin{equation}
    \bm{e}_q = \sum_{s=1}^{S} \mathbf{c}_{s,i_s}
\end{equation}
with its discrete representation expressed as the index sequence $[i_1, i_2, \dots, i_S]$. 
In our implementation, we empirically set $S=32$, which achieves an effective trade-off between reconstruction fidelity and sequence compactness.

To optimize the codebooks and quantized representations jointly, we adopt the standard VQ objective augmented with residual refinements. The overall loss is:  

\begin{equation}
\mathcal{L}_{\mathrm{rc}} \;=\;
\underbrace{\|\bm{e}-\bm{e}_q\|^2}_{\text{reconstruction loss}}
\;+\;
\underbrace{\sum_{s=1}^{S}\big\|\mathrm{sg}[\mathbf{res}_{s-1}]-\mathbf{c}_{s,i_s}\big\|^2}_{\text{codebook loss}}
\end{equation}

\begin{equation}
\mathcal{L}_{\mathrm{RVQ}} \;=\; 
\mathcal{L}_{\mathrm{rc}}
\;+\;
\beta \underbrace{\sum_{s=1}^{S}\big\|\mathbf{res}_{s-1}-\mathrm{sg}[\mathbf{c}_{s,i_s}]\big\|^2}_{\text{commitment loss}}
\end{equation}
where $\mathrm{sg}[\cdot]$ denotes the stop-gradient operator and $\beta$ is a weighting hyperparameter. The term $\mathcal{L}_{\mathrm{rc}}$ combines two objectives: the reconstruction loss, which enforces fidelity between the pretrained embedding $\mathbf{e}$ and its quantized approximation $\mathbf{e}_q$, and the codebook loss, which moves the codewords toward the residual vectors. The additional commitment term, weighted by $\beta$, prevents codebook collapse by encouraging each residual to remain close to its selected codeword.

Through the residual vector quantization process, the pretrained semantic embedding $\mathbf{e}$ is transformed into a compact discrete sequence of $S$ code indices. These indices not only preserve the semantic structure encoded by RotatE but also enable direct integration with LLM vocabularies via token mapping in subsequent stages.

\subsection{Fine-tuning LLMs with Quantized Tokens}

We extend the LLM vocabulary with $K$ new tokens ${[\texttt{CODE}_j]}$, each corresponding to a codeword in the residual vector quantization codebooks. Each codeword embedding is obtained by summing the vectors at the same position across all $S$ stages of the codebooks. Since the dimensionality of $\mathbf{c}_{j}^{\text{sum}}$ matches the LLM embedding size, we initialize the embedding of each new token with this sum, ensuring that it captures the full semantic information represented by the residual vector quantization codebooks:

\begin{equation}
    E([\texttt{CODE}_j]) = \sum_{s=1}^{S} \mathbf{c}_{s,j}
\end{equation}

During adaptation, two update strategies are applied in a complementary manner. First, the embeddings of the newly added tokens are trained with standard gradient descent, allowing them to refine beyond their initialization and better integrate into the LLM semantic space. The embeddings of the original vocabulary remain frozen to preserve the pretrained linguistic capacity.

Second, the internal parameters of the LLM are adapted using \textbf{Low-Rank Adaptation} \cite{LoRA}, which introduces trainable low-rank matrices into the attention and feed-forward layers:

\begin{equation}
\begin{gathered}
    W' = W + AB^\top,\\
    A \in \mathbb{R}^{d_\text{LLM} \times r}, \quad 
    B \in \mathbb{R}^{k_\text{LLM} \times r}, \quad 
    r \ll \min(d_\text{LLM}, k_\text{LLM})
\end{gathered}
\end{equation}
where $d_\text{LLM}$ and $k_\text{LLM}$ denote the input and output dimensions of the weight matrix $W$, respectively. LoRA introduces a low-rank adjustment $AB^\top$ to $W$, allowing efficient fine-tuning by training only $A$ and $B$ while keeping $W$ fixed.

This efficient adaptation enables the model to internalize the discrete quantized token format and align it with its generative process, thereby producing structured outputs that match the desired prediction format. The overall fine-tuning objective remains next-token prediction:

\begin{equation}
    \mathcal{L}_\text{LLM} = \sum_{n=1}^{N} \log p(x_n \mid x_{<n})
\end{equation}
where $x_n$ denotes the $n$-th token in the input sequence, which may consist of both natural language tokens and quantized KG tokens (the detailed construction of these mixed training sequences is described in Appendix A). Through this dual fine-tuning scheme, which combines direct optimization of new token embeddings with LoRA-based adaptation of internal parameters, the LLM learns to interpret and generate discrete knowledge graph representations while preserving its original linguistic proficiency.

\subsection{Ontology-Guided Class Integration}

In addition to entity-level quantization, we further integrate \textbf{ontology knowledge} by modeling entity classes. Each entity $e \in \mathcal{E}$ is associated with a class $y(e) \in \mathcal{Y}$, and $y(e)$ maps an entity to its corresponding class. This allows us to incorporate class-level semantic constraints into the KG completion process.

To integrate class information into our framework, we treat classes as entities and apply the same three-step procedure: (i) pretraining class embeddings with RotatE on the class graph, (ii) quantizing these embeddings using residual vector quantization, and (iii) extending the LLM vocabulary with class-specific tokens initialized from the quantized codewords. The LLM is then fine-tuned to predict missing classes in incomplete triples $(h,r,?)$, analogous to entity prediction.

Since the number of classes is significantly smaller than the number of entities, the LLM achieves \textbf{higher prediction accuracy on classes}. We exploit this property as a semantic constraint: given an entity prediction $\hat{t}$ for a query $(h,r,?)$, we simultaneously predict its class $\hat{y}$. If $\hat{t}$ does not belong to $\hat{y}$, we refine the prediction by selecting an alternative entity candidate $t'$ such that $y(t') = \hat{y}$. 


This ontology-guided filtering mechanism enforces \textbf{class consistency} between entity and class predictions, thereby improving robustness and semantic correctness in KG completion.

\section{Experiments}

In this section, we conduct a comprehensive empirical study to evaluate the proposed model ReaLM. The design of our experiments is guided by four central research questions:
\textbf{RQ1:} How well does the ReaLM model perform on the standard link prediction task?
\textbf{RQ2:} Can the ReaLM model be effectively extended to additional tasks such as triple classification?
\textbf{RQ3:} How well does the residual vector quantization preserve semantics in the learned representations?

\subsection{Experimental Setup}

\begin{table*}
  \caption{KG link prediction results of our model RealLM and the baselines including the traditional embedding-based and LLM-based models. All results of the baselines are sourced from the original papers. The best and second-best results are \textbf{boldfaced} and \underline{underlined}, respectively.
  }
  \label{tab:lp}
  \begin{tabular}{lcccccccc}
    \toprule
    \multirow{2}{*}{\parbox[c]{2cm}{\centering \textbf{Model}}} 
    & \multicolumn{4}{c}{\textbf{WN18RR}} 
    & \multicolumn{4}{c}{\textbf{FB15k237}} \\
    \cmidrule(lr){2-5} \cmidrule(lr){6-9}
    & \textbf{MRR} & \textbf{Hits@1} & \textbf{Hits@3} & \textbf{Hits@10} & \textbf{MRR} & \textbf{Hits@1} & \textbf{Hits@3} & \textbf{Hits@10} \\
    \midrule
    \rowcolor[gray]{0.9} \multicolumn{9}{c}{\textbf{Embedding Methods}} \\
    TransE \cite{TransE} & 0.223 & 0.014 & 0.401 & 0.529 & 0.330 & 0.231 & 0.369 & 0.528 \\
    RotatE \cite{RotatE} & 0.476 & 0.428 & 0.492 & 0.571 & 0.338 & 0.241 & 0.375 & 0.533 \\
    ConvE \cite{ConvE} & 0.460 & 0.390 & 0.430  & 0.480 & 0.316 & 0.239 & 0.350 & 0.491 \\
    HyConvE \cite{HyConvE} & 0.461 & 0.432 & - & 0.534 & 0.339 & 0.212 & - & 0.458 \\
    HittER \cite{HittER} & 0.503 & 0.462 & 0.516 & 0.584 & 0.373 & 0.279 & 0.409 & 0.558 \\
    TCRA \cite{TCRA} & 0.496 & 0.457 & 0.511 & 0.574 & 0.367 & 0.275 & 0.403 & 0.554 \\
    \midrule
    \rowcolor[gray]{0.9} \multicolumn{9}{c}{\textbf{LLM-based Methods}} \\
    KG-BERT \cite{KG-BERT} & 0.216 & 0.041 & 0.302 & 0.524 & - & - & - & 0.520 \\
    KICGPT \cite{KICGPT} & 0.549 & 0.474 & 0.585 & 0.641 & 0.412 & 0.327 & 0.448 & 0.554 \\
    KG-FIT \cite{KG-FIT} & 0.553 & 0.488 & 0.595 & \underline{0.695} & 0.362 & 0.275 & 0.485 & 0.572 \\
    MKGL \cite{MKGL} & 0.552 & 0.500 & 0.577 & 0.656 & 0.415 & 0.325 & 0.454 & 0.591 \\
    SSQR-LLaMA2 \cite{SSQR} & 0.591 & 0.548 & \textbf{0.618} & 0.673 & 0.449 & 0.374 & \underline{0.491} & \underline{0.597} \\
    SSQR-LLaMA3.1 \cite{SSQR} & \underline{0.598} & \underline{0.559} & \textbf{0.618} & 0.675 & \underline{0.459} & \underline{0.393} & \underline{0.491} & \underline{0.597} \\
    \midrule
    ReaLM (Our Method) & \textbf{0.608} & \textbf{0.575} & \underline{0.615} & \textbf{0.699} & \textbf{0.467} & \textbf{0.402} & \textbf{0.498} & \textbf{0.603} \\
    \bottomrule
  \end{tabular}
\end{table*}

To systematically evaluate the performance and generalization capability of the proposed ReaLM, we conduct experiments on two representative KG reasoning tasks: \textit{link prediction} and \textit{triple classification}. We employ two widely used benchmark datasets, FB15k-237 \cite{FB15K237} and WN18RR \cite{WN18RR}, along with their ontology-augmented counterparts, FB15k-237O and WN18RRO \cite{OL-KGC}. To obtain semantic embeddings of entities and relations, we adopt the RotatE model, which has demonstrated strong performance in capturing relational patterns. Both entity and relation embeddings are pretrained with a dimensionality of 2048, aligned with the token embedding size of the base LLM to facilitate seamless integration.

The semantic embeddings are discretized using residual vector quantization. We configure each codebook with 1000 codewords, and each entity is represented by a sequence of 32 code indices, balancing reconstruction fidelity and compactness. For adaptation to LLMs, we extend the vocabulary of Llama-3.2 with quantized code tokens. Fine-tuning is conducted in two complementary ways: (i) optimizing the embeddings of the newly introduced tokens, and (ii) applying parameter-efficient LoRA adaptation to the internal layers of model. The maximum training epoch is set to 100.

For link prediction, we report mean reciprocal rank (MRR) and Hits@$k$, which are standard metrics for evaluating ranking-based tasks. For triple classification, we adopt accuracy, precision, recall, and F1-score, providing a holistic view of classification effectiveness.

\subsection{Link Prediction Analysis}

For link prediction experiments, we evaluated a range of representative KG embedding methods, including TransE \cite{TransE}, RotatE \cite{RotatE}, ConvE \cite{ConvE}, HyConvE \cite{HyConvE}, HittER \cite{HittER}, and TCRA \cite{TCRA}, alongside recent LLM-based approaches such as KICGPT \cite{KICGPT}, KG-FIT \cite{KG-FIT}, MKGL \cite{MKGL}, and SSQR \cite{SSQR}. All models were tested on the standard benchmark datasets WN18RR and FB15k237. The ontology information was extracted from the WN18RRO and FB15K237O datasets \cite{OL-KGC}.

\textbf{Result Analysis. }Table~\ref{tab:lp} presents the link prediction results on WN18RR and FB15k237. Traditional embedding methods, such as TransE and ConvE, achieve moderate performance, while recent LLM-based models provide noticeable improvements due to their contextual reasoning capabilities. Our proposed ReaLM consistently outperforms all baselines across both datasets. On WN18RR, ReaLM achieves an MRR of 0.608 and Hits@10 of 0.699, improving over the strongest LLM-based baseline by 1.6–3.5\%, with particularly strong gains in ranking the most relevant entities (Hits@1 of 0.575). On FB15k237, ReaLM attains an MRR of 0.467 and Hits@10 of 0.603, surpassing prior methods in all metrics and demonstrating robust generalization across datasets. Overall, these results indicate that ReaLM consistently outperforms both embedding-based and existing LLM-based approaches. This improvement can be attributed to its multi-stage design: high-quality RotatE embeddings encode rich relational structure, residual vector quantization transforms them into compact, semantically informative tokens, and the LLM is fine-tuned with LoRA to internalize these structured representations while maintaining linguistic proficiency. By bridging structured KG knowledge and the generative capabilities of LLMs, ReaLM achieves superior link prediction performance across datasets.

\begin{table}[h]
  \centering
  \caption{Comparison of link prediction and ablation experiment performance on WN18RR and FB15k-237 datasets. }
    \label{tab:ablation}
    \begin{tabular}{lcccc}
    \toprule
    \textbf{Model} & \textbf{MRR} & \textbf{Hits@1} & \textbf{Hits@3} & \textbf{Hits@10} \\
    \midrule
    \multicolumn{5}{c}{\cellcolor[gray]{0.9}\textbf{WN18RR}} \\
    ReaLM & 0.608 & 0.575 & 0.615 & 0.699 \\
    $\Delta$ ($\uparrow$) & 1.64\% & \cellcolor[HTML]{FFC0CB}2.78\% & -0.49\% & \cellcolor[HTML]{FFC0CB}3.43\% \\
    w/o ontology & 0.591 & 0.562 & 0.598 & 0.689 \\
    $\Delta$ ($\downarrow$) & \cellcolor[HTML]{FFC0CB}2.80\% & \cellcolor[HTML]{FFC0CB}2.26\% & \cellcolor[HTML]{FFC0CB}2.76\% & 1.43\% \\
    \midrule
    \multicolumn{5}{c}{\cellcolor[gray]{0.9}\textbf{FB15k-237}} \\
    ReaLM & 0.467 & 0.402 & 0.498 & 0.603 \\
    $\Delta$ ($\uparrow$) & 1.71\% & \cellcolor[HTML]{FFC0CB}2.24\% & 1.41\% & 1.00\%\\
    w/o ontology & 0.452 & 0.382 & 0.476 & 0.588 \\
    $\Delta$ ($\downarrow$) & \cellcolor[HTML]{FFC0CB}3.21\% & \cellcolor[HTML]{FFC0CB}4.98\% & \cellcolor[HTML]{FFC0CB}4.42\% & \cellcolor[HTML]{FFC0CB}2.49\% \\
    \bottomrule
    \end{tabular}
\end{table}

\textbf{Ablation Study. }To further investigate the contribution of ontology knowledge, we conducted ablation experiments in which the ontology-guided class integration component of ReaLM was removed. All other configurations, including entity embedding with RotatE, residual vector quantization, and LoRA-based LLM fine-tuning, were kept identical to the main experiments to isolate the effect of ontology information.

As shown in Table~\ref{tab:ablation}, the $\Delta$ symbols indicate performance changes with respect to the baseline SSQR model ($\uparrow$ for improvement, $\downarrow$ for degradation). The cells highlighted in pink denote statistically significant changes. Removing ontology knowledge leads to consistent performance degradation across both WN18RR and FB15k237. On WN18RR, MRR decreases from 0.608 to 0.591, Hits@1 drops sharply from 0.575 to 0.562, and Hits@10 declines moderately from 0.699 to 0.689. Similarly, on FB15k237, MRR falls from 0.467 to 0.452, Hits@1 from 0.402 to 0.382, and Hits@10 from 0.603 to 0.588.

Notably, the pronounced drop in Hits@1 highlights that ontology knowledge is particularly critical for top-rank prediction accuracy, ensuring that the most relevant entities are correctly identified. The smaller reductions in Hits@10 suggest that overall ranking quality is less affected, but class-level constraints still provide a measurable boost in precision. These observations confirm that integrating ontology information enables ReaLM to maintain semantic consistency, improving both the reliability and robustness of link prediction across different datasets.

\begin{figure}[h]
  \centering
  \begin{subfigure}[b]{0.45\linewidth}
    \includegraphics[width=\linewidth]{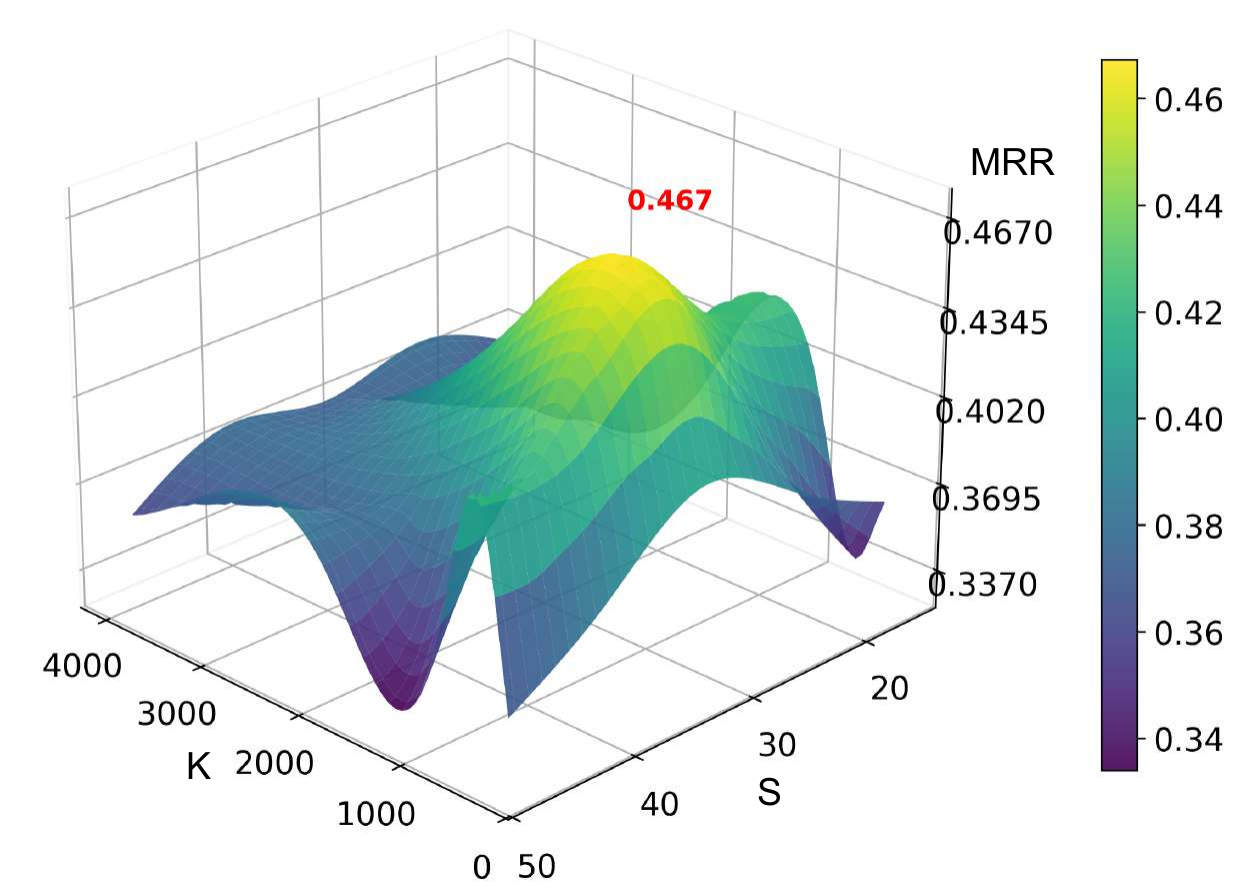}
    \caption{FB15k-237 Dataset}
    \label{fig:FB15k-237_3D}
  \end{subfigure}
  \hfill 
  \begin{subfigure}[b]{0.45\linewidth}
    \includegraphics[width=\linewidth]{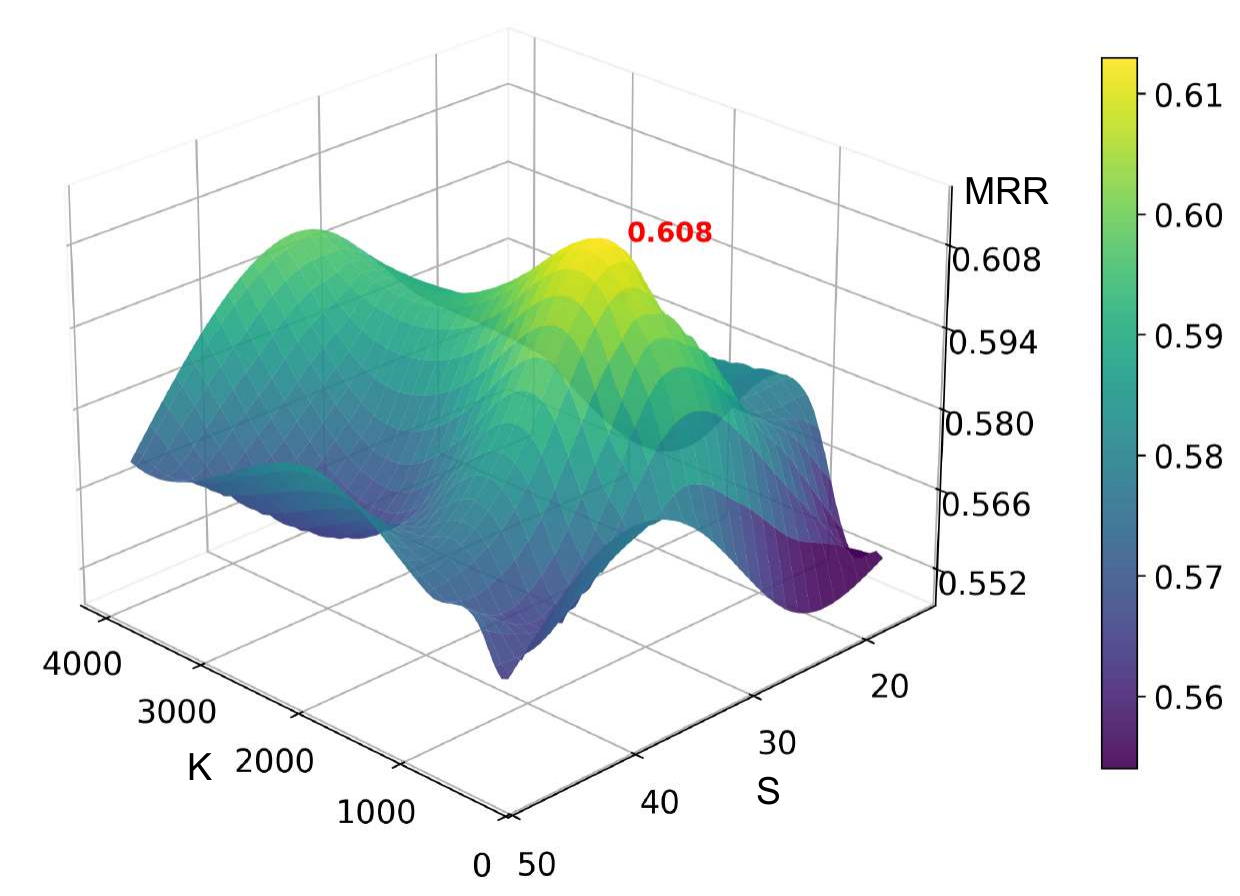}
    \caption{WN18RR Dataset}
    \label{fig:WN18RR_3D}
  \end{subfigure}
  \caption{Impact of codebook size and quantization stages on link prediction performance}
  \label{fig:3D}
\end{figure}

\textbf{Analysis of Residual Vector Quantization Parameters.}
To investigate the impact of residual vector quantization hyperparameters on performance of ReaLM, we conducted a grid search over codebook size $K$ and the number of quantization stages $S$. Figures~\ref{fig:3D}(a) and (b) present 3D surface plots of link prediction MRR scores on FB15k-237 and WN18RR, respectively.

The results indicate that both $K$ and $S$ significantly influence model performance. On FB15k-237, the MRR achieves its peak (0.467) at $K=1000$, $S=32$, suggesting that moderate codebook sizes and a sufficient number of stages are critical for capturing rich semantic information while avoiding over-compression. On WN18RR, a similar pattern emerges, with the highest MRR (0.608) observed at $K=1500$ and $S=32$. Performance tends to degrade when $K$ is too small, likely due to insufficient representational capacity, or when $S$ is excessively large, possibly introducing redundancy and overfitting in the discrete token representations. Overall, this analysis demonstrates that careful tuning of the residual vector quantization parameters enables ReaLM to produce compact yet semantically informative token sequences, thereby maximizing the effectiveness of LLM fine-tuning for link prediction tasks \textbf{(answering RQ1)}.

\subsection{Triple Classification Analysis}

\textbf{Result Analysis. }As presented in Table \ref{tab:TC_A}, ReaLM achieves the highest performance among all compared models, reaching an accuracy of 0.814 and an F1-score of 0.825 on the FB15k-237N dataset. Compared with the strongest baseline SSQR-LLaMA3.1, ReaLM improves the F1-score by 1.4 percentage points, demonstrating its superior ability to distinguish valid triples from corrupted ones.
The improvement is consistent across all metrics, with gains of 1.7\% in accuracy, 1.7\% in precision, and 1.4\% in recall. This balanced enhancement indicates that ReaLM not only identifies a wider range of correct triples but also maintains high discriminative precision. Such robustness suggests that the model effectively captures both global relational structures and local semantic constraints.

\begin{table}[h]
  \centering
  \caption{The experiment results of the triple classification on FB15k-237N dataset. The results of baselines are taken from SSQR \cite{SSQR}.}
  \label{tab:TC_A}
    \begin{tabular}{lcccc}
    \toprule
    \textbf{Model} & \cellcolor{gray!25}\textbf{Acc} & \textbf{P} & \textbf{R} & \cellcolor{gray!25}\textbf{F1} \\
    \midrule
    TransE \cite{TransE} & \cellcolor{gray!25}0.697 & 0.708 & 0.671 & \cellcolor{gray!25}0.689 \\
    ComplEx \cite{ComplEx} & \cellcolor{gray!25}65.70 & 66.46 & 63.38 & \cellcolor{gray!25}64.88 \\
    DistMult \cite{DistMult} & \cellcolor{gray!25}0.587 & 0.590 & 0.568 & \cellcolor{gray!25}0.579 \\
    RotatE \cite{RotatE} & \cellcolor{gray!25}0.684 & 0.692 & 0.664 & \cellcolor{gray!25}0.678 \\
    \midrule
    Alpaca & \cellcolor{gray!25}0.561 & 0.533 & 0.974 & \cellcolor{gray!25}0.689 \\
    GPT-3.5 & \cellcolor{gray!25}0.602 & 0.866 & 0.240 & \cellcolor{gray!25}0.376 \\
    KG-LLaMA \cite{KG-LLaMA} & \cellcolor{gray!25}0.748 & 0.674 & 0.962 & \cellcolor{gray!25}0.793 \\
    KG-Alpaca \cite{KG-LLaMA} & \cellcolor{gray!25}0.699 & 0.627 & 0.983 & \cellcolor{gray!25}0.766 \\
    KoPA \cite{KoPA} & \cellcolor{gray!25}0.777 & 0.708 & 0.941 & \cellcolor{gray!25}0.808 \\
    SSQR-LLaMA2 \cite{SSQR} & \cellcolor{gray!25}0.794 & 0.757 & 0.867 & \cellcolor{gray!25}0.808 \\
    SSQR-LLaMA3.1 \cite{SSQR} & \cellcolor{gray!25}0.798 & 0.759 & 0.872 & \cellcolor{gray!25}0.811 \\
    \midrule
    ReaLM & \cellcolor{gray!25}0.814 & 0.776 & 0.886 & \cellcolor{gray!25}0.825 \\
    w/o ontology & \cellcolor{gray!25}0.783 & 0.754 & 0.856 & \cellcolor{gray!25}0.798 \\
    $\Delta$($\downarrow$) & \cellcolor{gray!25}3.96\% & 2.92\% & 3.50\% & \cellcolor{gray!25}3.38\% \\
    \bottomrule
    \end{tabular}
\end{table}

The strong performance of ReaLM can be attributed to its design that integrates discrete KG representations into the token space of LLM through residual vector quantization. This alignment allows the language model to internalize entity-level semantics while retaining symbolic interpretability. Consequently, ReaLM establishes a tighter semantic correspondence between KG embeddings and textual representations, leading to more consistent reasoning and reliable triple classification.

\textbf{Ablation Study on Ontology Knowledge. }To examine the contribution of ontology-guided semantic constraints, we conduct an ablation study by removing the ontology integration module from ReaLM. In this setting, entities are still quantized and integrated into the LLM vocabulary, but the model no longer models class-level structures. Specifically, we omit the three-step process of pretraining, quantizing, and tokenizing class embeddings, and the LLM is trained solely to predict entities without incorporating class-level supervision. As a result, the model lacks the ontology-guided refinement mechanism that enforces semantic consistency between predicted entities and their corresponding classes.

As shown in Table \ref{tab:TC_A}, removing ontology knowledge leads to a consistent decline across all metrics, with decreases of 3.96\% in accuracy and 3.38\% in F1-score. This degradation indicates that, without class-level constraints, the model becomes more prone to semantic drift — predicting entities that are contextually plausible but type-inconsistent with the relational pattern. In other words, while the quantized entity embeddings preserve relational structure, the absence of ontology guidance weakens the ability of ReaLM to filter out type-incoherent predictions.
The results demonstrate that ontology-guided filtering plays a critical role in refining predictions of ReaLM. By aligning entity and class predictions, it effectively constrains the search space to semantically valid candidates, improving both precision and recall.

\begin{figure}[h]
  \centering
  \includegraphics[width=\linewidth]{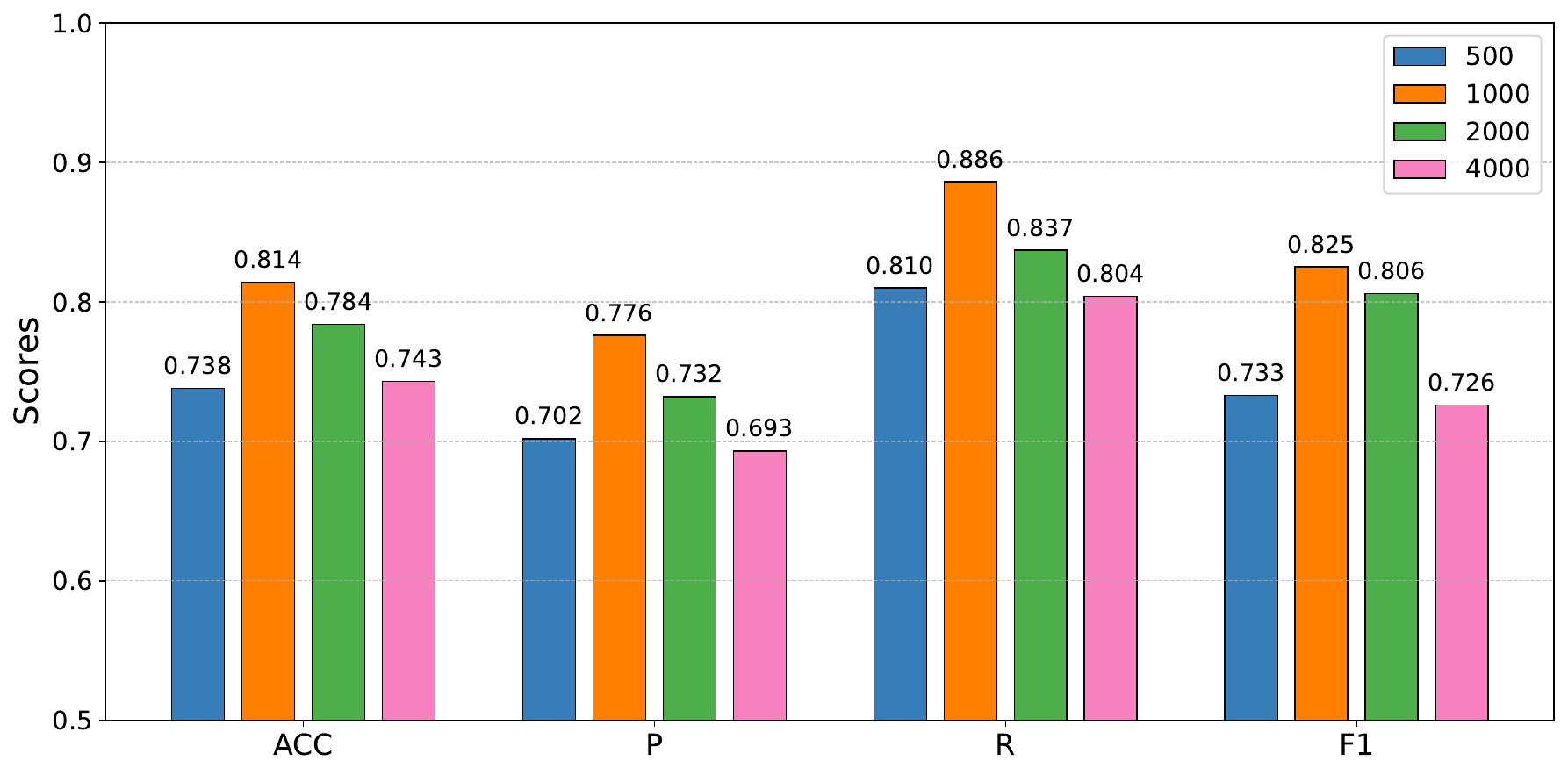}
  \caption{Impact of codebook size on triple-classification performance.}
  \Description{Bar chart showing the impact of codebook size on triple-classification performance metrics (ACC, P, R, F1) for different codebook sizes on the FB15k-237 dataset.}
  \label{fig:codebook_size}
\end{figure}

\textbf{Effect of Codebook Size.} We evaluate the influence of the residual vector quantization codebook size \(K\) on triple-classification performance. In our setup, each new token \([\texttt{CODE}_j]\) is initialized by summing its stage-wise codewords as $E([\texttt{CODE}_j]) = \sum_{s=1}^{S} \mathbf{c}_{s,j}$, 
where \(\mathbf{c}_{s,j}\) denotes the \(j\)-th codeword in stage \(s\). The embeddings of these added tokens are trainable while the original LLM vocabulary remains frozen, and the internal parameters of the model are adapted via LoRA. We report accuracy, precision, recall, and F1 for \(K=\{500,1000,2000,4000\}\) in Figure \ref{fig:codebook_size}.

Two clear patterns emerge from the results. First, increasing $K$ from small to moderate values yields consistent performance gains: a larger codebook improves representational granularity and allows more distinct quantized codes to capture subtle entity differences, which in turn helps the model discriminate valid from corrupted triples. Second, improvement saturates beyond an intermediate scale: further enlarging $K$ produces diminishing returns and may introduce marginal instability. 
This saturation effect can be explained by two complementary factors. On one hand, a larger $K$ increases expressiveness (finer quantization) and therefore tends to raise recall by enabling the model to represent more valid candidates distinctly. On the other hand, when $K$ becomes very large relative to the effective coverage of the training data, many codewords are infrequently activated; sparse updates to their initialized embeddings reduce their reliability and can slightly degrade precision or produce noisy distinctions.

From a practical standpoint on the FB15k-237N dataset, an intermediate codebook size offers the best trade-off between expressiveness, trainability, and computational cost. The dual adaptation scheme employed in this work, which combines trainable new-token embeddings with low-rank LoRA updates, helps mitigate sparsity issues by enabling contextual integration of quantized tokens without requiring updates to the entire model. Nevertheless, overly large $K$ still increases memory usage and the number of task-specific parameters, and requires more data or stronger regularization to avoid overfitting to rare codewords \textbf{(answering RQ2)}.

\subsection{Quantitative Code Analysis}

To examine the impact of the codebook configuration on quantization performance, we varied the number of codewords $K \in {4,8,16,32,64}$ and evaluated three indicators: reconstruction mean squared error , commitment MSE, and cosine similarity. The reconstruction and commitment MSE respectively measure how accurately the quantized embeddings approximate the original representations and how well the codewords fit the latent distribution, while cosine similarity reflects the semantic alignment between the quantized and continuous embeddings.

\begin{figure}[h]
  \centering
  \includegraphics[width=0.8\linewidth]{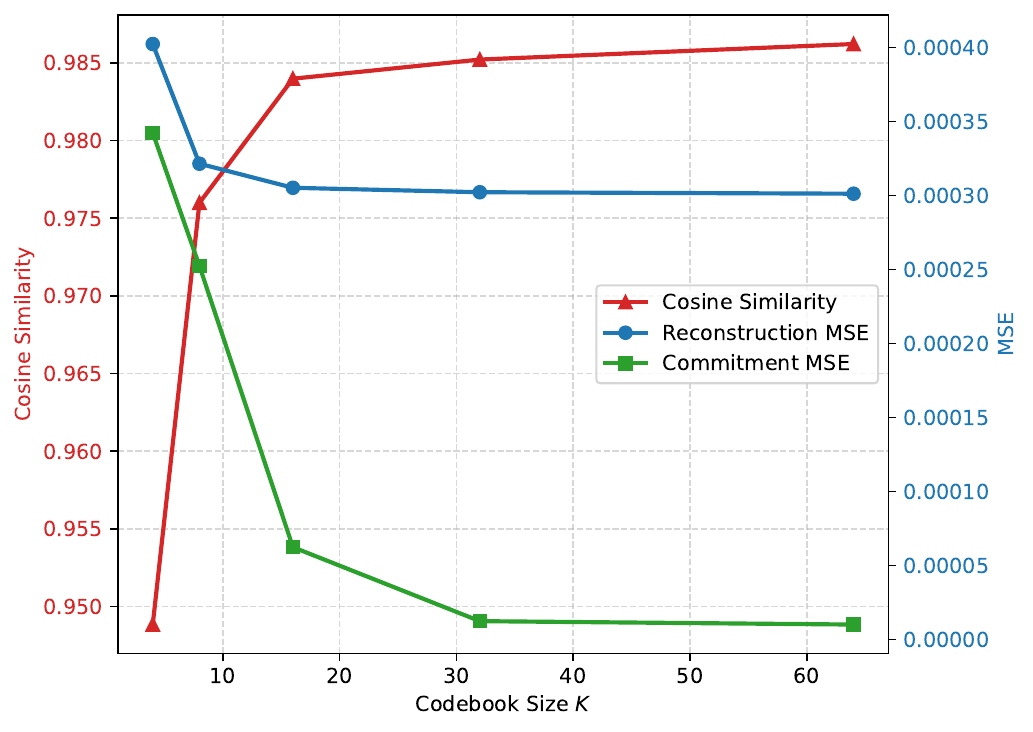}
  \caption{Impact of codebook size on triple-classification performance.}
  \Description{Bar chart showing the impact of codebook size on triple-classification performance metrics (ACC, P, R, F1) for different codebook sizes on the FB15k-237 dataset.}
  \label{fig:QVC-2}
\end{figure}

As shown in Figure \ref{fig:QVC-2}, increasing the codebook size consistently reduces both reconstruction and commitment errors while improving cosine similarity. Specifically, the reconstruction MSE decreases from 0.000402 to 0.000301, and the commitment MSE drops by nearly two orders of magnitude, whereas the cosine similarity rises from 0.9488 to 0.9862. These results indicate that enlarging the codebook enhances the ability of model to preserve both numerical precision and semantic consistency. Overall, the proposed quantization scheme effectively compresses the KG embeddings while maintaining their relational semantics, demonstrating strong representational efficiency \textbf{(answering RQ3)}.

\section{Conclusion}

In this paper, we proposed \textbf{ReaLM} that integrates knowledge graph semantics into large language models through residual vector quantization and ontology-guided refinement. By discretizing high-dimensional KG embeddings into compact code sequences, ReaLM effectively bridges the gap between continuous entity representations and the discrete token space of LLMs. Extensive experiments on standard benchmarks demonstrate that ReaLM achieves state-of-the-art performance across both link prediction and triple classification tasks, confirming its capability to internalize and reason over structured knowledge while maintaining general language understanding.

\bibliographystyle{ACM-Reference-Format}
\bibliography{main.bib}

@inproceedings{efficient,
  title={An efficient parallel keyword search engine on knowledge graphs},
  author={Yang, Yueji and Agrawal, Divykant and Jagadish, HV and Tung, Anthony KH and Wu, Shuang},
  booktitle={2019 IEEE 35th International Conference on Data Engineering},
  pages={338--349},
  year={2019},
  organization={IEEE}
}

@inproceedings{learning,
  title={Learning intents behind interactions with knowledge graph for recommendation},
  author={Wang, Xiang and Huang, Tinglin and Wang, Dingxian and Yuan, Yancheng and Liu, Zhenguang and He, Xiangnan and Chua, Tat-Seng},
  booktitle={Proceedings of the Web Conference 2021},
  pages={878--887},
  year={2021}
}

@inproceedings{beyond,
  title={Beyond iid: three levels of generalization for question answering on knowledge bases},
  author={Gu, Yu and Kase, Sue and Vanni, Michelle and Sadler, Brian and Liang, Percy and Yan, Xifeng and Su, Yu},
  booktitle={Proceedings of the Web Conference 2021},
  pages={3477--3488},
  year={2021}
}

@inproceedings{TransE,
  title={Translating embeddings for modeling multi-relational data},
  author={Bordes, Antoine and Usunier, Nicolas and Garcia-Dur{\'a}n, Alberto and Weston, Jason and Yakhnenko, Oksana},
  booktitle={Proceedings of the 26th International Conference on Neural Information Processing Systems},
  pages={2787--2795},
  year={2013},
  series = {NeurIPS'13}
}

@inproceedings{TransH,
  title={Knowledge Graph Embedding by Translating on Hyperplanes},
  author={Wang, Zhen and Zhang, Jianwen and Feng, Jianlin and Chen, Zheng},
  booktitle={Proceedings of the Twenty-Eighth AAAI Conference on Artificial Intelligence},
  pages={1112--1119},
  year={2014},
  series = {AAAI'14}
}

@inproceedings{TransD,
  title={Knowledge graph embedding via dynamic mapping matrix},
  author={Ji, Guoliang and He, Shizhu and Xu, Liheng and Liu, Kang and Zhao, Jun},
  booktitle={Proceedings of the 53rd Annual Meeting of the Association for Computational Linguistics and the 7th International Joint Conference on Natural Language Processing},
  pages={687--696},
  year={2015},
  series = {ACL'15}
}

@inproceedings{TransM,
  title={Transition-based knowledge graph embedding with relational mapping properties},
  author={Fan, Miao and Zhou, Qiang and Chang, Emily and Zheng, Fang},
  booktitle={Proceedings of the 28th Pacific Asia Conference on Language, Information and Computing},
  pages={328--337},
  year={2014}
}

@inproceedings{5E,
  title={5* knowledge graph embeddings with projective transformations},
  author={Nayyeri, Mojtaba and Vahdati, Sahar and Aykul, Can and Lehmann, Jens},
  booktitle={Proceedings of the AAAI Conference on Artificial Intelligence},
  volume={35},
  number={10},
  pages={9064--9072},
  year={2021}
}

@inproceedings{DistMult,
  title = {Embedding Entities and Relations for Learning and Inference in Knowledge Bases},
  author = {Yang, Bishan and Yih, Scott Wen-tau and He, Xiaodong and Gao, Jianfeng and Deng, Li},
  booktitle = {Proceedings of the third International Conference on Learning Representations},
  year = {2015},
  pages = {809--816},
  series = {ICLR'15}
}

@inproceedings{ComplEx,
  title={Complex embeddings for simple link prediction},
  author={Trouillon, Th{\'e}o and Welbl, Johannes and Riedel, Sebastian and Gaussier, {\'E}ric and Bouchard, Guillaume},
  booktitle={Proceedings of the 33rd International Conference on International Conference on Machine Learning},
  pages={2071--2080},
  year={2016},
  series = {ICML'16}
}

@inproceedings{RotatE,
  title={RotatE: Knowledge Graph Embedding by Relational Rotation in Complex Space},
  author={Zhiqing Sun and Zhi-Hong Deng and Jian-Yun Nie and Jian Tang},
  booktitle={Proceedings of the Seventh International Conference on Learning Representations},
  pages={328--337},
  year={2019},
  series = {ICLR'19}
}

@inproceedings{ConvE,
    title = {Convolutional 2D Knowledge Graph Embeddings},
    author = {Dettmers, Tim and Minervini, Pasquale and Stenetorp, Pontus and Riedel, Sebastian},
    booktitle = {Proceedings of the Thirty-Second AAAI Conference on Artificial Intelligence},
    pages = {1811--1818},
    year = {2018},
    series = {AAAI'18}
}

@inproceedings{ConvKB,
    title = {A Novel Embedding Model for Knowledge Base Completion Based on Convolutional Neural Network},
    author = {Nguyen, Dai Quoc and Nguyen, Tu Dinh and Nguyen, Dat Quoc and Phung, Dinh},
    booktitle = {Proceedings of the 2018 Conference of the North American Chapter of the Association for Computational Linguistics: Human Language Technologies},
    pages = {327--333},
    year = {2018},
    series = {NAACL'18}
}

@inproceedings{ConvR,
  title={Adaptive convolution for multi-relational learning},
  author={Jiang, Xiaotian and Wang, Quan and Wang, Bin},
  booktitle={Proceedings of the 2019 Conference of the North American Chapter of the Association for Computational Linguistics: Human Language Technologies, Volume 1 (Long and Short Papers)},
  pages={978--987},
  year={2019},
  series = {NAACL'20}
}

@article{RelAtt,
  title={Knowledge graph embedding using graph convolutional networks with relation-aware attention},
  author={Sheikh, Nasrullah and Qin, Xiao and Reinwald, Berthold and Miksovic, Christoph and Gschwind, Thomas and Scotton, Paolo},
  journal={arXiv preprint arXiv:2102.07200},
  year={2021}
}

@article{HRAN,
  title={Learning knowledge graph embedding with heterogeneous relation attention networks},
  author={Li, Zhifei and Liu, Hai and Zhang, Zhaoli and Liu, Tingting and Xiong, Neal N},
  journal={IEEE Transactions on Neural Networks and Learning Systems},
  volume={33},
  number={8},
  pages={3961--3973},
  year={2021},
  publisher={IEEE}
}

@inproceedings{KGT5,
  title={Sequence-to-Sequence Knowledge Graph Completion and Question Answering},
  author={Saxena, Apoorv and Kochsiek, Adrian and Gemulla, Rainer},
  booktitle={Proceedings of the 60th Annual Meeting of the Association for Computational Linguistics},
  pages={2814--2828},
  year={2022},
  series = {ACL'22}
}

@inproceedings{HittER,
  title={HittER: Hierarchical Transformers for Knowledge Graph Embeddings},
  author={Chen, Sanxing and Liu, Xiaodong and Gao, Jianfeng and Jiao, Jian and Zhang, Ruofei and Ji, Yangfeng},
  booktitle={Proceedings of the 2021 Conference on Empirical Methods in Natural Language Processing},
  pages={10395--10407},
  year={2021},
  series = {EMNLP'21}
}

@inproceedings{HyConvE,
  title={HyConvE: A Novel Embedding Model for Knowledge Hypergraph Link Prediction with Convolutional Neural Networks},
  author={Wang, Chenxu and Wang, Xin and Li, Zhao and Chen, Zirui and Li, Jianxin},
  booktitle={Proceedings of the ACM Web Conference 2023},
  pages={188--198},
  year={2023},
  series = {WWW'23}
}

@inproceedings{FB15K237,
  title={Observed versus latent features for knowledge base and text inference},
  author={Toutanova, Kristina and Chen, Danqi},
  booktitle={Proceedings of the 3rd Workshop on Continuous Vector Space Models and their Compositionality},
  year={2015},
  series = {ACL'15}
}

@inproceedings{WN18RR,
  title={Convolutional 2d knowledge graph embeddings},
  author={Dettmers, Tim and Minervini, Pasquale and Stenetorp, Pontus and Riedel, Sebastian},
  booktitle={Proceedings of the AAAI Conference on Artificial Intelligence},
  year={2018},
  series = {AAAI'18}
}

@article{ConvD,
  title={Convd: Attention enhanced dynamic convolutional embeddings for knowledge graph completion},
  author={Guo, Wenbin and Li, Zhao and Wang, Xin and Chen, Zirui and Zhao, Jun and Li, Jianxin and Yuan, Ye},
  journal={IEEE Transactions on Knowledge and Data Engineering},
  year={2025},
  publisher={IEEE}
}

@inproceedings{OL-KGC,
  title={Ontology-Enhanced Knowledge Graph Completion using Large Language Models},
  author={Guo, Wenbin and Wang, Xin and Chen, Jiaoyan and Li, Zhao and Chen, Zirui},
  booktitle={THE 24th INTERNATIONAL SEMANTIC WEB CONFERENCE (ISWC)},
  year={2025}
}

@article{MSHE,
  title={Multisource hierarchical neural network for knowledge graph embedding},
  author={Jiang, Dan and Wang, Ronggui and Xue, Lixia and Yang, Juan},
  journal={Expert Systems with Applications},
  volume={237},
  pages={121446},
  year={2024},
  publisher={Elsevier}
}

@inproceedings{KGEditor,
  title={Editing language model-based knowledge graph embeddings},
  author={Cheng, Siyuan and Zhang, Ningyu and Tian, Bozhong and Chen, Xi and Liu, Qingbin and Chen, Huajun},
  booktitle={Proceedings of the AAAI Conference on Artificial Intelligence},
  volume={38},
  number={16},
  pages={17835--17843},
  year={2024}
}

@inproceedings{summar20241,
  author={Feng, Tuoyu and Wu, Yongsheng and Li, Libing},
  booktitle={2024 9th International Conference on Computer and Communication Systems (ICCCS)}, 
  title={Research on Knowledge Graph Completion Based Upon Knowledge Graph Embedding}, 
  year={2024},
  volume={},
  number={},
  pages={1335-1342},
  doi={10.1109/ICCCS61882.2024.10603073}}

@Article{KLR-KGC,
AUTHOR = {Ji, Shengwei and Liu, Longfei and Xi, Jizhong and Zhang, Xiaoxue and Li, Xinlu},
TITLE = {KLR-KGC: Knowledge-Guided LLM Reasoning for Knowledge Graph Completion},
JOURNAL = {Electronics},
VOLUME = {13},
YEAR = {2024},
NUMBER = {24},
ARTICLE-NUMBER = {5037},
URL = {https://www.mdpi.com/2079-9292/13/24/5037},
ISSN = {2079-9292},
DOI = {10.3390/electronics13245037}
}

@article{TEA-GLM,
  title={LLMs as Zero-shot Graph Learners: Alignment of GNN Representations with LLM Token Embeddings},
  author={Wang, Duo and Zuo, Yuan and Li, Fengzhi and Wu, Junjie},
  journal={arXiv preprint arXiv:2408.14512},
  year={2024}
}

@inproceedings{KoPA,
  title={Making large language models perform better in knowledge graph completion},
  author={Zhang, Yichi and Chen, Zhuo and Guo, Lingbing and Xu, Yajing and Zhang, Wen and Chen, Huajun},
  booktitle={Proceedings of the 32nd ACM International Conference on Multimedia},
  pages={233--242},
  year={2024}
}

@article{KG-BERT,
  title={KG-BERT: BERT for Knowledge Graph Completion},
  author={Yao, Liang and Mao, Chengsheng and Luo, Yuan},
  journal={arXiv e-prints},
  pages={arXiv--1909},
  year={2019}
}

@article{KG-LLaMA,
  title={Exploring large language models for knowledge graph completion},
  author={Yao, Liang and Peng, Jiazhen and Mao, Chengsheng and Luo, Yuan},
  journal={arXiv preprint arXiv:2308.13916},
  year={2023}
}

@inproceedings{MuKDC-1,
  title={LLM-based multi-level knowledge generation for few-shot knowledge graph completion},
  author={Li, Qian and Chen, Zhuo and Ji, Cheng and Jiang, Shiqi and Li, Jianxin},
  booktitle={Proceedings of the Thirty-Third International Joint Conference on Artificial Intelligence},
  volume={271494703},
  year={2024}
}

@inproceedings{KnowC,
  title={Knowledge Context Modeling with Pre-trained Language Models for Contrastive Knowledge Graph Completion},
  author={Yang, Guangqian and Liu, Yi and Zhang, Lei and Zhang, Licheng and Xie, Hongtao and Mao, Zhendong},
  booktitle={Findings of the Association for Computational Linguistics ACL 2024},
  pages={8619--8630},
  year={2024}
}

@inproceedings{TCRA,
  title={A unified joint approach with topological context learning and rule augmentation for knowledge graph completion},
  author={Guo, Jingtao and Zhang, Chunxia and Li, Lingxi and Xue, Xiaojun and Niu, Zhendong},
  booktitle={Findings of the association for computational linguistics ACL 2024},
  pages={13686--13696},
  year={2024}
}

@article{KG-FIT,
  title={Kg-fit: Knowledge graph fine-tuning upon open-world knowledge},
  author={Jiang, Pengcheng and Cao, Lang and Xiao, Cao Danica and Bhatia, Parminder and Sun, Jimeng and Han, Jiawei},
  journal={Advances in Neural Information Processing Systems},
  volume={37},
  pages={136220--136258},
  year={2024}
}

@inproceedings{KICGPT,
  title={KICGPT: Large Language Model with Knowledge in Context for Knowledge Graph Completion},
  author={Wei, Yanbin and Huang, Qiushi and Zhang, Yu and Kwok, James},
  booktitle={Findings of the Association for Computational Linguistics: EMNLP 2023},
  pages={8667--8683},
  year={2023}
}

@inproceedings{LoRA,
title={Lo{RA}: Low-Rank Adaptation of Large Language Models},
author={Edward J Hu and Yelong Shen and Phillip Wallis and Zeyuan Allen-Zhu and Yuanzhi Li and Shean Wang and Lu Wang and Weizhu Chen},
booktitle={International Conference on Learning Representations},
year={2022},
url={https://openreview.net/forum?id=nZeVKeeFYf9}
}

@article{MKGL,
  title={MKGL: mastery of a three-word language},
  author={Guo, Lingbing and Bo, Zhongpu and Chen, Zhuo and Zhang, Yichi and Chen, Jiaoyan and Yarong, Lan and Sun, Mengshu and Zhang, Zhiqiang and Luo, Yangyifei and Li, Qian and others},
  journal={Advances in Neural Information Processing Systems},
  volume={37},
  pages={140509--140534},
  year={2024}
}

@inproceedings{Galkin2022,
  title={NodePiece: Compositional and Parameter-Efficient Representations of Large Knowledge Graphs},
  author={Galkin, Mikhail and Denis, Etienne and Wu, Jiapeng and Hamilton, William L},
  booktitle={International Conference on Learning Representations},
  year={2022}
}

@inproceedings{Chen2023,
  title={Entity-agnostic representation learning for parameter-efficient knowledge graph embedding},
  author={Chen, Mingyang and Zhang, Wen and Yao, Zhen and Zhu, Yushan and Gao, Yang and Pan, Jeff Z and Chen, Huajun},
  booktitle={Proceedings of the AAAI conference on artificial intelligence},
  volume={37},
  number={4},
  pages={4182--4190},
  year={2023}
}

@inproceedings{Li2023,
  title={Random Entity Quantization for Parameter-Efficient Compositional Knowledge Graph Representation},
  author={Li, Jiaang and Wang, Quan and Liu, Yi and Zhang, Licheng and Mao, Zhendong},
  booktitle={The 2023 Conference on Empirical Methods in Natural Language Processing},
  year={2023}
}

@article{Li2021,
  title={Discrete Knowledge Graph Embedding based on Discrete Optimization},
  author={Li, Yunqi and Xu, Shuyuan and Liu, Bo and Fu, Zuohui and Liu, Shuchang and Chen, Xu and Zhang, Yongfeng},
  journal={arXiv e-prints},
  pages={arXiv--2101},
  year={2021}
}

@inproceedings{Zhu2022,
  title={Dualde: Dually distilling knowledge graph embedding for faster and cheaper reasoning},
  author={Zhu, Yushan and Zhang, Wen and Chen, Mingyang and Chen, Hui and Cheng, Xu and Zhang, Wei and Chen, Huajun},
  booktitle={Proceedings of the Fifteenth ACM International Conference on Web Search and Data Mining},
  pages={1516--1524},
  year={2022}
}

@article{SSQR,
  author       = {Qika Lin and
                  Tianzhe Zhao and
                  Kai He and
                  Zhen Peng and
                  Fangzhi Xu and
                  Ling Huang and
                  Jingying Ma and
                  Mengling Feng},
  title        = {Self-supervised Quantized Representation for Seamlessly Integrating Knowledge Graphs with Large Language Models},
  booktitle    = {Proceedings of the 63st Annual Meeting of the Association for Computational Linguistics (ACL)},
  year         = {2025}
}

\end{document}